\documentclass[conference]{IEEEtran} 
\IEEEoverridecommandlockouts
\usepackage{cite}
\usepackage{amsmath,amssymb,amsfonts}
\usepackage{algorithm}
\usepackage{algorithmic}
\usepackage{graphicx}
\usepackage{textcomp}
\usepackage{xcolor}
\usepackage[numbers]{natbib}
\usepackage{booktabs} 
\usepackage{todonotes}
\usepackage{tabularx}
\usepackage{url}
\usepackage{acronym}
\usepackage{natbib}
\usepackage{amsmath}
\usepackage{paralist}
\usepackage{balance}
\newcounter{matriz}

\usepackage{paralist}
\usepackage{fancyvrb}
\usepackage{comment}
\usepackage{cleveref}
\usepackage{array}
\usepackage{caption} 
\usepackage{refcount}

\usepackage[detect-all]{siunitx}
\newcolumntype{P}[1]{>{\centering\arraybackslash}p{#1}}
\usepackage[inline, shortlabels]{enumitem}
\usepackage{listings}
\acrodef{dais}[DAIS-ITA]{International Technology Alliance for Distributed Analytics and Information Sciences}
\acrodef{gpbm}[GPM]{Generative Policy-based Model}
\acrodef{ce}[CE]{ITA Controlled English}
\def\BibTeX{{\rm B\kern-.05em{\sc i\kern-.025em b}\kern-.08em
    T\kern-.1667em\lower.7ex\hbox{E}\kern-.125emX}}
\begin{document}

\title{Synthetic Ground Truth Generation for\\Evaluating Generative Policy Models}

\author{\IEEEauthorblockN{Daniel Cunnington}
\IEEEauthorblockA{\textit{Emerging Technology} \\
\textit{IBM Research}\\
Hursley, UK \\
dancunnington@uk.ibm.com}
\and
\IEEEauthorblockN{Graham White}
\IEEEauthorblockA{\textit{Emerging Technology} \\
\textit{IBM Research}\\
Hursley, UK \\
gwhite@uk.ibm.com}
\and
\IEEEauthorblockN{Geeth de Mel}
\IEEEauthorblockA{\textit{Hartree Centre} \\
\textit{IBM Research}\\
Daresbury, UK \\
geeth.demel@uk.ibm.com}
}

\maketitle

\begin{abstract}
\acp{gpbm} aim to enable a coalition of systems---be they devices or services---to adapt according to contextual changes such as environmental factors, user preferences and different tasks whilst adhering to various constraints and regulations as directed by a managing party or the collective vision of the coalition. Recent developments have proposed new architectures to realize the potential of~\acp{gpbm} but as the complexity of systems and their associated requirements increases, there is an emerging requirement to have scenarios and associated datasets to realistically evaluate~\acp{gpbm} with respect to the properties of the operating environment---be it the future battlespace or an autonomous organization. In order to address this requirement, in this paper, we present a method of applying an agile knowledge representation framework to model requirements---both individualistic and collective---that enables synthetic generation of ground truth data such that advanced~\acp{gpbm} can be evaluated robustly in complex environments. We also release conceptual models, annotated datasets, as well as means to extend the data generation approach so that similar datasets can be developed for varying complexities and different situations.
\end{abstract}


\section{Introduction}

Within the \acf{dais}\footnote{\url{https://dais-ita.org/}\label{dais-pub}}, fundamental research is being performed to realise the future battlespace envisioned by various military Doctrine papers~\cite{army_values_2018, mod_2018}. The \ac{dais} programme focuses on two technical areas; \begin{inparaenum}[(1)]
\item \textit{Dynamic Secure Coalition Information Infrastructures} and
\item \textit{Coalition Distributed Analytics and Situational Understanding}\end{inparaenum}. In the future battlespace, coalition systems and devices will be required to operate in challenging environments that impose certain constraints such as lack of or low bandwidth connectivity to backend services, rapidly changing environmental conditions and the requirement to abide by legal regulations and mission directives. Therefore, coalition systems and devices require the capability to adapt and evolve such that they can behave autonomously `at the edge' in previously unseen contexts. Crucially, systems need to understand the bounds in which they can operate based on their own (and that of other systems) capability, constraints of the environment and safety requirements. Also, systems and devices with varying autonomous capability may be required to collaborate with humans and other coalition partners to achieve a shared coalition goal, thus outlining the requirement for explainable actions and decisions. Recent work within the \ac{dais} programme has developed the notion of \acfp{gpbm} for use in coalition environments~\cite{Verma:2017gen} to enable coalition systems to define their behaviour and adapt to new situations whilst conforming to a variety of constraints and regulations. 

In order to evaluate our work and demonstrate it's applicability to coalition environments, illustrative scenarios and accompanying datasets are required. In a military scenario, this is challenging due to the lack of publicly available datasets and mission-specific information. Current efforts to date utilize non-military scenarios and datasets, such as open source traffic cameras for congestion classification~\cite{willis2017deep} and IoT related deployments for smart-home policy management~\cite{goynugur2017knowledge}. Also, recent work has outlined an approach for generating synthetic data based on a Connected and Autonomous Vehicle (CAV) scenario~\cite{CunningtonVTC2018, Cunnington2019ICDCS} utilising an agile knowledge representation framework that supports human-machine conversations as well as reasoning and hypothesis testing.

This paper extends the high level \ac{dais} scenario~\cite{white2019dais} and develops the knowledge representation approach for synthetic data generation by enabling the user to control the complexity of the dataset to support future evaluations as the fundamental research progresses. The paper is structured as follows. In Section~\ref{sec:background} we discuss a set of illustrative vignettes to motivate our work alongside relevant background knowledge. In Section ~\ref{sec:knowledge_rep} we outline the knowledge representation approach including \ac{dais} specific domain models and discuss the approach for generating data with varying complexity. Finally, we conclude the paper in Section~\ref{sec:conclusion} by providing closing remarks.

\section{Background}\label{sec:background}
This section outlines a set of background knowledge including two military focused vignettes that may occur in a coalition operation, based on the high-level \ac{dais} scenario~\cite{white2019dais}. In the proposed vignettes, let us assume three partners are working together in a coalition, namely the UK, US and a fictitious Non-Government Organization `Kish' in order to achieve a mission objective in a given region. In this section we also discuss a potential challenge for coalitions surrounding asset serviceability as a future application for generative policy research.

\subsection{Person of Interest Tracking}
\begin{figure}[!h]
\centering
\includegraphics[width=\columnwidth]{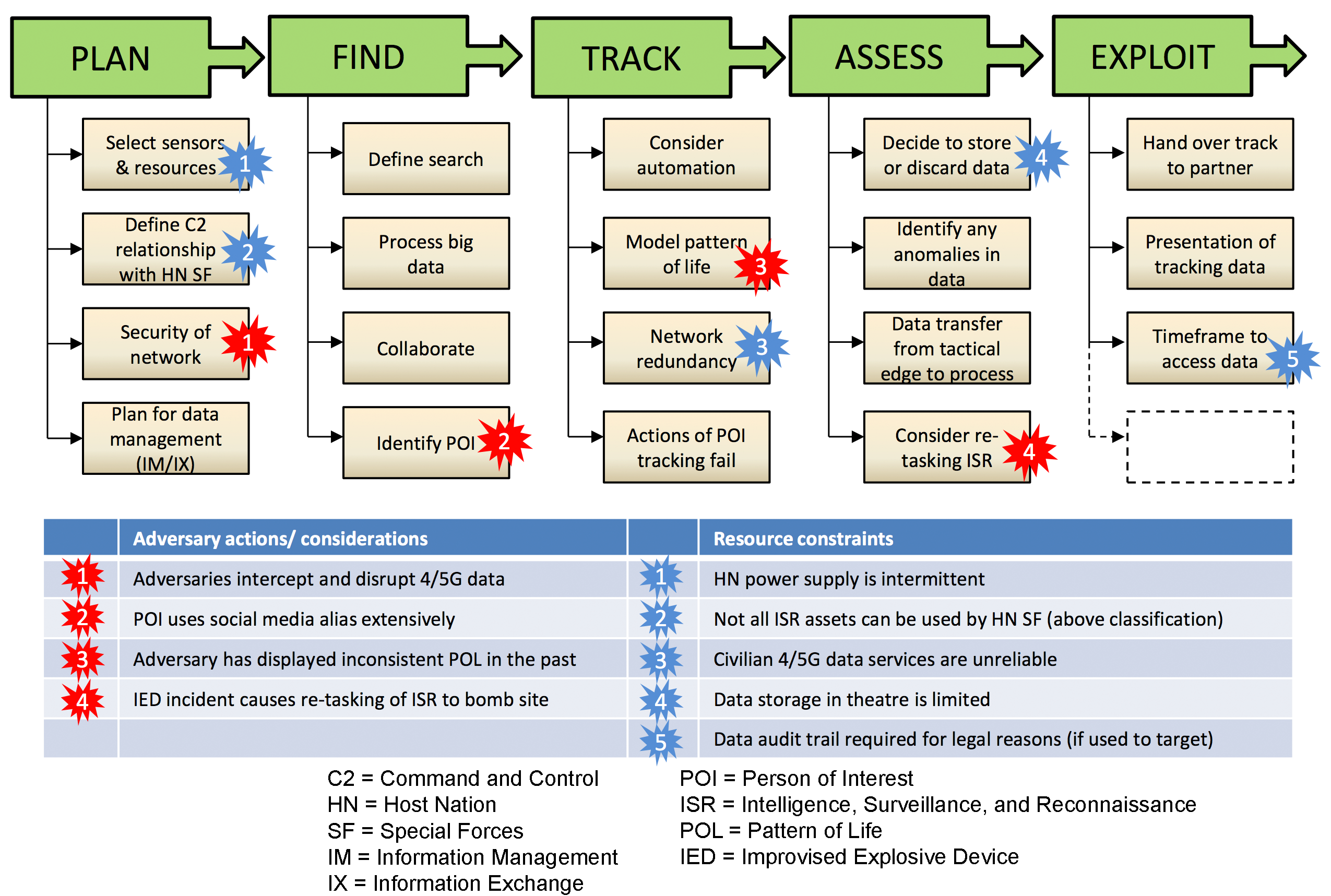}
\caption{Person of Interest Tracking Mission, obtained from ~\cite{white2019dais}}
\label{fig:poi_tracking}
\end{figure}

As detailed in \cite{white2019dais}, one possible mission the coalition could be tasked with is `Person of Interest Tracking' which involves the monitoring of a High-Value Target (HVT). A detailed breakdown of the various phases of activity within this mission is shown in Figure~\ref{fig:poi_tracking} including potential adversary actions and resource constraints.

In this scenario, different coalition partners will have different roles and responsibilities. For example, UK forces could partner with Kish's local security force to execute the `Plan' and `Find' stages and the US may partner with the UK and Kish for the `Track', `Assess' and `Exploit' stages. The planning stage involves identifying a set of coalition assets and resources to use which may be constrained by power and connectivity, as well as being vulnerable to man-in-the-middle attacks from an adversary intercepting network communications. Then, the coalition finds and tracks the person of interest, and finally collates the results. Throughout the mission, coalition assets are subject to re-tasking, e.g. if the enemy plants an Improvised Explosive Device (IED), coalition forces and assets may have to relocate to deal with the IED. Also, the coalition has to manage data storage, power and network connectivity constraints dynamically throughout the mission.

\subsection{Logistical Resupply}
\begin{figure}[!h]
\centering
\includegraphics[width=\columnwidth]{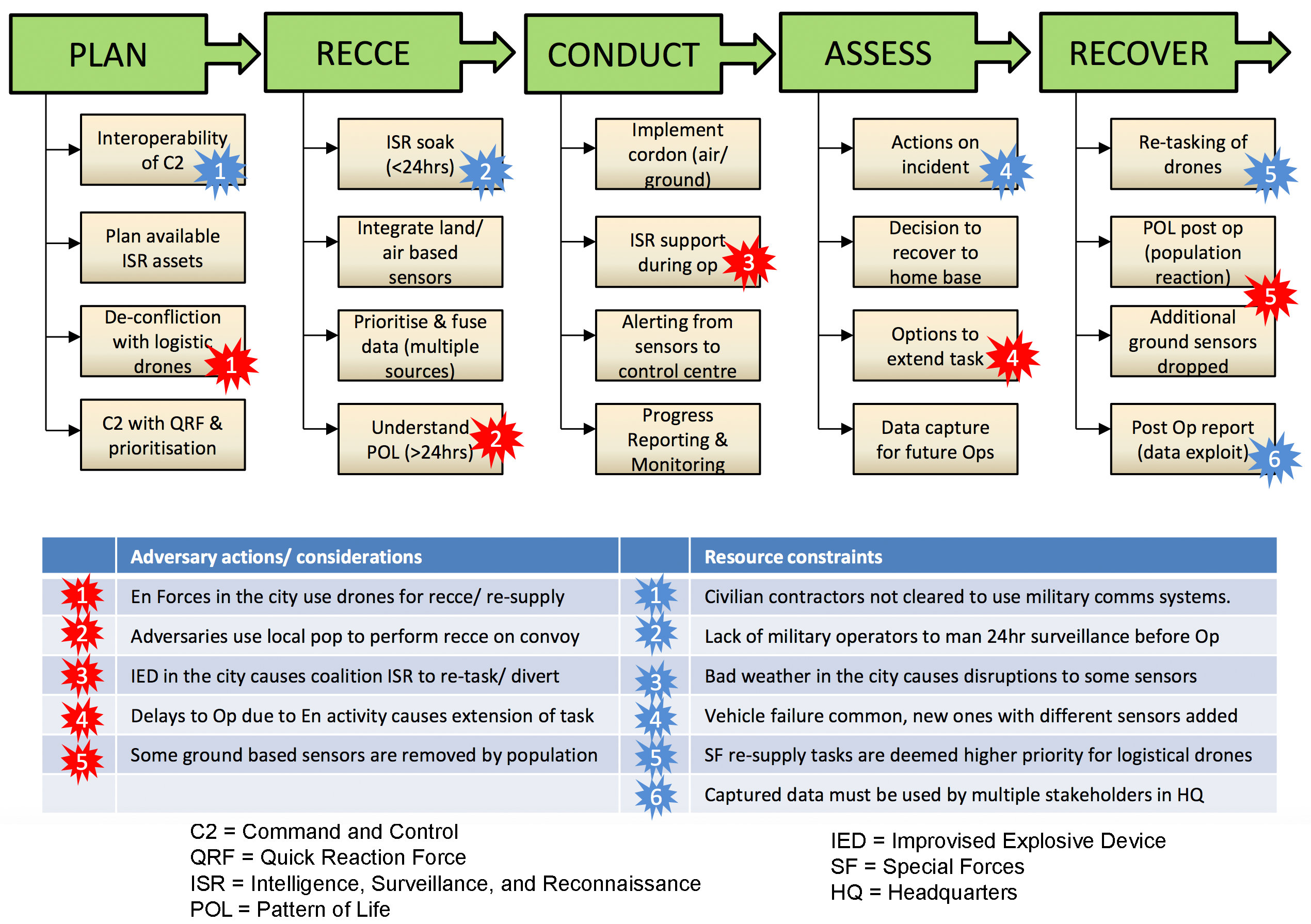}
\caption{Logistical Resupply Mission, obtained from ~\cite{white2019dais}}
\label{fig:resupply}
\end{figure}

The second vignette describes a logistical resupply mission to serve coalition troops that are stationed in a given location such as a city. A convoy is provided by a civilian contractor and is accompanied by a guarding force, or force
protection element, from another coalition partner. In addition, regular resupply of troops in the city is conducted by autonomous resupply drones, that also have other tasks such as resupply of high priority special forces elements in mountainous locations outside the city. Similarly to the first vignette, this vignette is
broken down into phases of activity as shown in Figure~\ref{fig:resupply}. These take into account elements
such as the planning phase where intelligence, surveillance and reconnaissance assets will have to be assigned to the mission; monitoring of the mission during its execution; and the recovery phase back to base.
The enemy will once again be employing disruption tactics and the coalition is constrained by the resources available for use.

\subsection{Military Environments}

As detailed in~\cite{white2019dais}, the missions outlined above could potentially be executed in a wide range of environments, which impose various constraints on the types of activities that can occur for a given mission. This is detailed in Table~\ref{table:mis}.
\begin{table}[!ht] 
  \centering
  \renewcommand{\arraystretch}{1.5}
  \begin{tabular}{|P{2.3cm}|P{1cm}|P{1cm}|P{1cm}|P{1.3cm}|}
    \cline{2-5}
    \multicolumn{1}{c|}{} & \textbf{Urban} & \textbf{Desert} & \textbf{Jungle} & \textbf{Mountain} \\ \hline
    Number of non-combatants & High  & Low & Low & Low \\ \cline{1-5}
    Amount of valuable infrastructure & High  & Low & Low & Low \\ \cline{1-5}
    Presence of multi-dimensional battlespace & Yes  & No & Some & Yes \\ \cline{1-5}
    Restricted rules of engagement & Yes  & No & Some & Yes \\ \cline{1-5}
    Detection, observation, engagement ranges & Short  & Long & Short & Medium \\ \cline{1-5}
    Avenues of approach & Many  & Many & Few & Few \\ \cline{1-5}
    Freedom of movement \& manoeuvre & Low  & High & Low & Medium \\ \cline{1-5}
    Communications Functionality & Less  & Normal & Normal & Less \\ \cline{1-5}
    Logistical Requirements & High  & High & Medium & Medium \\ \cline{1-5}
    \end{tabular}
\caption{\label{table:mis}Possible Mission Environments, obtained from ~\cite{white2019dais}}
\end{table}

\subsection{ALFUS Framework: Autonomy Levels for Unmanned Systems}
\label{sec:background:alfus}
\begin{figure}[!h]
\centering
\includegraphics[width=\columnwidth]{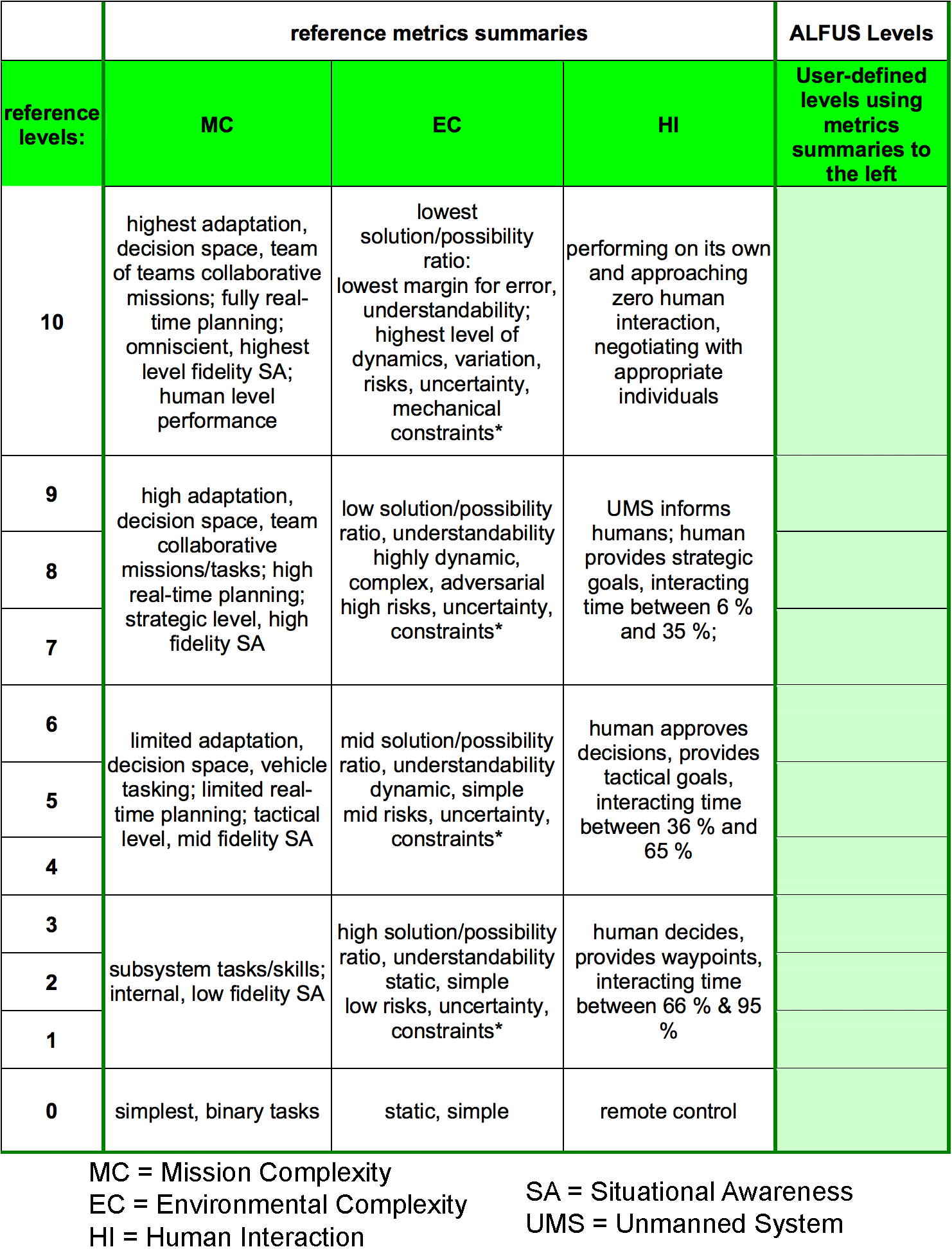}
\caption{Autonomy Levels for Unmanned Systems (ALFUS), obtained from ~\cite{huang2007autonomy}}
\label{fig:alfus}
\end{figure}

Huang has outlined varying levels of autonomy, ranging from \numrange[range-phrase = --]{0}{10} inclusive that describe the full range of autonomous capability in unmanned systems~\cite{huang2007autonomy}. Figure~\ref{fig:alfus} details Mission Complexity (MC), Environmental Complexity (EC) and Human Interaction (HI) capabilities at each level. The overall ALFUS level is obtained through scoring each capability with a value \numrange[range-phrase = --]{0}{3} inclusive, for levels \numrange[range-phrase = --]{1}{9}. At level 0, each capability receives a 0 score to indicate no autonomous capability and level 10 remains independent to represent the ultimate goal for autonomy.

\subsection{Asset Serviceability}
In order to perform the two missions outlined in Figures~\ref{fig:poi_tracking} and~\ref{fig:resupply}, coalition partners are required to work together which may involve sharing assets with one another. An asset could be a physical device such as a surveillance camera, an autonomous system such as an unmanned aerial vehicle or an asset could be a virtual service such as a database or a specific type of machine learning model such as a face recognizer or weaponry detector. Each asset has a set of time-bound constraints that govern it’s use at a given point in time. Possible constraint types are defined as follows: \begin{inparaenum}[\itshape(1)]
\item Physical Constraints---the asset can only operate in certain physical locations or contexts. Also, in order to utilize an asset the requestor may need to be within a suitable physical range of the asset;
\item Security Constraints---certain functionality within the asset may be withheld by the asset owner through access control. For example, the high value targets database may only accept read requests from a requestor iff they are requesting information about a given target.
\item Trust Constraints---certain assets or functionality may only be shared between coalition partners iff the trust value between partners is above a predefined threshold\end{inparaenum}. In the above definitions, constraints can be local to the individual asset in a given context (e.g. a certain autonomous vehicle is currently unable to drive autonomously given high wind speeds) or constraints can be applied at the global level by the coalition partner (e.g. do not share the facial recognition model unless the trust with the requesting partner is greater than or equal to a predefined threshold). Given the constantly evolving context and potentially dynamic asset pool caused by coalition partners frequently bringing assets online/offline during a mission, local and global serviceability policies that define the type of requests that an asset or coalition partner can service will be constantly in flux and it won’t be possible for a human operator to generate all of these policies for all assets in all possible contexts in a timely manner. Also, given the distributed nature of a partner’s assets, the coalition partner as a whole requires means to keep track of the serviceability of its asset pool through a generated policy and also is responsible for communicating this updated policy in a distributed manner to its assets. For example, a US autonomous vehicle, denoted \emph{Vehicle A} moves to a new location in the city so that the camera and object detector on-board the vehicle can now fulfill requests for information regarding a new type of object (e.g. civilian pedestrians are now in view). \emph{Vehicle A} now needs to generate a new local policy that states that it can service these new requests and this policy needs to be communicated to all US assets, so that the US can generate a global policy that states that if any asset within the US’ inventory receives a request regarding the street \emph{Vehicle A} is located at, the asset can direct the requestor to \emph{Vehicle A}, or, utilize the Software Defined Network to perform the request to \emph{Vehicle A} on behalf of the requestor. 

\section{Approach}\label{sec:knowledge_rep}
Utilizing the existing knowledge and literature presented in Section~\ref{sec:background}, as well as the requirement to account for asset serviceability, in this section we discuss our approach for representing the coalition environment by creating \ac{dais} specific domain models to support synthetic data generation.

\subsection{Knowledge Representation}

\ac{ce}~\cite{braines2013controlled} was chosen as a semantic knowledge base due to it's expressive power and
built-in reasoning capabilities that enable future human-machine conversational interfaces to be developed that utilize the conceptual model. Also, \ac{ce} enables domain models to be easily specified using human-readable, controlled natural language which ensures our work is extendable if necessary. Figure~\ref{fig:ce-missions} details the concept definitions for a \emph{mission} and a \emph{mission instance} in \ac{ce}. This is inspired by the illustrative vignettes outlined in Figures~\ref{fig:poi_tracking} and~\ref{fig:resupply} as missions contain high level stages, can be disrupted by potential adversary actions and are bound by certain constraints. Also, inspired by Table~\ref{table:mis}, missions can be executed in different mission environments and different environmental conditions, as well as being undertaken by a certain coalition at a certain date/time. Note that the \emph{coalition}, \emph{mission environment} and \emph{environmental condition instance} concepts are defined separately and are not included in Figure~\ref{fig:ce-missions} due to space limitations. We release the full concept definition online for further reference\footnote{\label{dais}\url{https://github.com/dais-ita/coalition-data}}.  

\begin{figure}
    \centering
    {\footnotesize{
\begin{lstlisting}
conceptualise a ~ mission ~ M that
  has the value 'S' as ~ high level stage ~ and
  has the value 'A' as ~ potential adversary action ~ and
  has the value 'C' as ~ constraint ~.

conceptualise a ~ mission instance ~ MI that
  ~ is an instance of ~ the mission M and
  ~ is executed by ~ the coalition C and
  ~ is executed in ~ the mission environment E and
  ~ is executed in ~ the environmental condition instance ECI and
  has the value 'T' as ~ start time ~.
\end{lstlisting}}}
    \caption{CE concept definitions for a mission and mission instance}
    \label{fig:ce-missions}
\end{figure}

\begin{figure}[!h]
\centering
\includegraphics[width=\columnwidth]{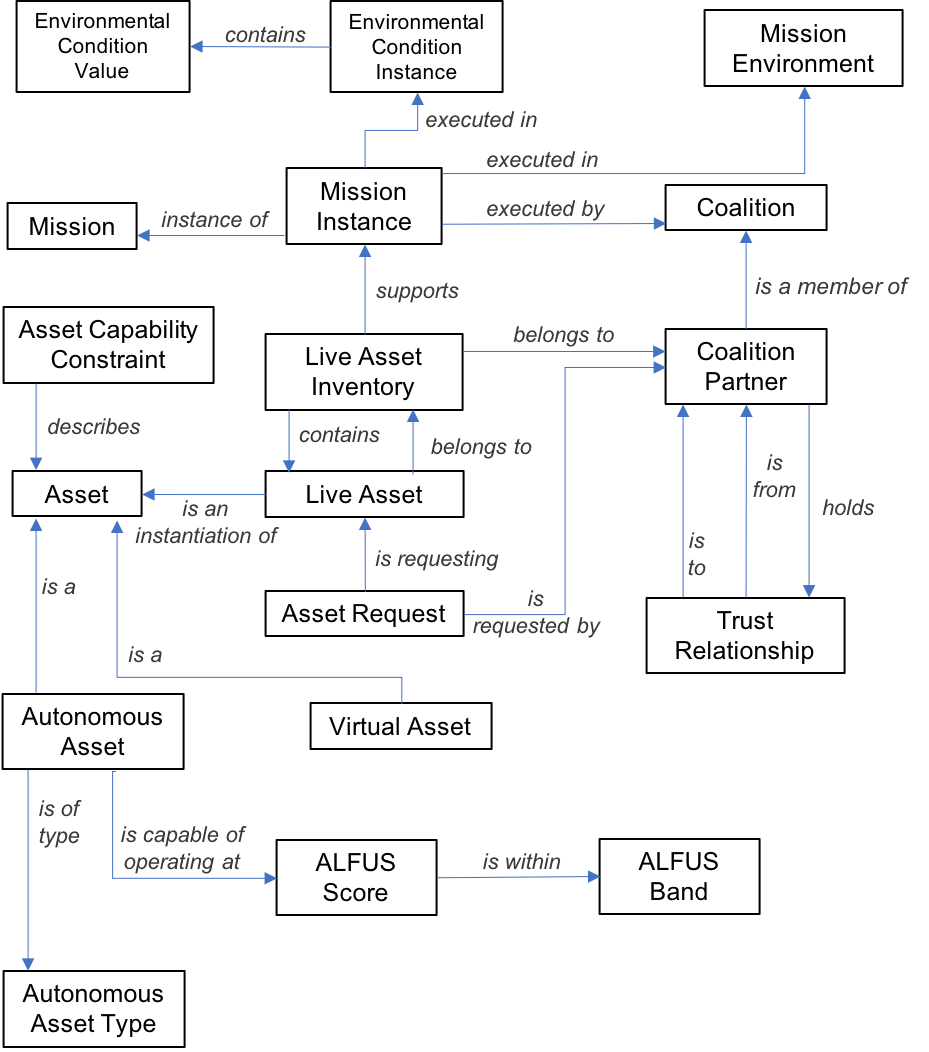}
\caption{CE Model illustrating concepts and relationships}
\label{fig:ce-model}
\end{figure}

The concept of a \emph{mission} and a \emph{mission instance} can then be instantiated in \ac{ce} as facts and stored in the knowledge base. For example, Figure~\ref{fig:ce-missions-facts} defines the person of interest tracking mission and instantiates it with respect to a coalition, mission environment, environmental conditions and a start time. Domain modelling is repeated for other concepts including \emph{ALFUS levels}, \emph{coalition partners}, \emph{coalition trust relationships}, \emph{environmental conditions} and \emph{assets} and stored in the knowledge base for instantiation. The full model illustrating concepts and their relationships is shown in Figure~\ref{fig:ce-model}.
\begin{figure}
    \centering
    {\footnotesize{
\begin{lstlisting}
there is a mission named 'person of interest tracking' that
  has the value 'plan' as high level stage and
  has the value 'find' as high level stage and
  has the value '4G/5G communication disruption' as potential adversary action and
  has the value 'POI uses social media alias extensively' as potential adversary action and
  has the value 'Limited data storage in theatre' as constraint and
  has the value 'Data Audit trail required for legal reasons' as constraint. 

there is a mission instance named 'mi_1' that
  is an instance of the mission 'person of interest tracking' and
  is executed by the coalition 'US/UK/KISH' and
  is executed in the mission environment 'urban' and
  is executed in the environmental condition instance 'eci_1' and
  has the value '2019-02-21 13:20' as start time.
\end{lstlisting}}}
    \caption{CE concept definitions for a mission and mission instance}
    \label{fig:ce-missions-facts}
\end{figure}

\subsection{Fact Generation}
Facts such as those detailed in Figure~\ref{fig:ce-missions-facts} can be used to generate a dataset that is required for evaluating \acp{gpbm}. However, in this scenario, there are a large amount of facts required in the knowledge base in order to represent a real-world coalition environment and it would be too cumbersome to generate these facts manually. Therefore, we define a small set of base concepts and generate facts programmatically as shown with the generalized approach in Algorithm~\ref{alg:facts}.

\begin{algorithm}
\caption{Generalized Fact Generation}
\label{alg:facts}
\begin{algorithmic}[1] 
\STATE $B \leftarrow$ base concept
\STATE $V \leftarrow$ \textit{generateValues($B$)}
\STATE $C \leftarrow$ \textit{generateAllCombinations($V$)}
\STATE $facts \leftarrow$ set()
\FOR{$c_{i}$ in $C$} 
    \STATE $CE \leftarrow$ \textit{generateCE($c_{i}$)} 
    \STATE $facts \leftarrow facts\cup CE$
\ENDFOR
\STATE return facts
\end{algorithmic}
\end{algorithm}

Firstly, we define a set of possible environmental conditions: \emph{visibility level}, \emph{temperature level}, \emph{rainfall level}, \emph{snowfall level}, \emph{wind speed level} and \emph{humidity level} where each condition contains an associated lower and upper bound, appropriate units and a weighting to denote the importance of this condition. This refers to line 1 in Algorithm~\ref{alg:facts}. \emph{Environmental condition values} are then generated for each condition between the lower and upper bounds according to a user-configurable granularity parameter---i.e. if the humidity level has a lower bound of 0\%, an upper bound of 100\% and the granularity parameter is 5, values 0\%, 25\%, 50\%, 75\% and 100\% would be generated. This refers to line 2 in Algorithm~\ref{alg:facts}. To ensure coverage of a wide range of possible conditions, an \emph{environmental condition instance} is generated for each possible combination of the environmental conditions---i.e. setting the granularity parameter to 5 and using the 6 different environmental conditions there would be $5^6=15,625$ different \emph{environmental condition instances} which corresponds to line 3 in Algorithm~\ref{alg:facts}. Also, during the combination process, we normalize each condition and compute a weighted average to denote the severity of each \emph{environmental condition instance}, according to the importance weight specified for each condition. Then we iterate over all the possible combinations and build a \ac{ce} fact sentence, such as those shown in Figure~\ref{fig:ce-missions-facts}. This corresponds to lines \numrange[range-phrase = --]{5}{8} in Algorithm~\ref{alg:facts}. Next, all the possible ALFUS scores are generated. Recall from Section~\ref{sec:background:alfus} that an overall ALFUS score is the sum of the capability scores, which identifies the level of complexity of a mission an asset can deal with, the level of environmental complexity an asset can operate in and the amount of human involvement an asset requires. Given each capability can be scored between \numrange[range-phrase = --]{0}{3} inclusive and that there are 3 capabilities, we generate $4^3=64$ overall ALFUS scores, representing all possible combinations of scores for mission complexity, environment complexity and human interaction. We manually add an additional score for ALFUS level 10 as level 10 is independent from other ALFUS levels. Given all \emph{coalitions}, \emph{mission environments}, \emph{mission types}, generated \emph {environmental condition instances} and a predefined list of 4 start times, we generate all possible combinations of \emph{mission instances} which ensures our dataset covers all possible conditions a mission could be executed under. Setting the granularity parameter for the environmental conditions to 5 and assuming there is 1 coalition, 4 mission environment types (Table~\ref{table:mis}) and 2 mission types, we generate $1*4*2*15625=125,000$ possible mission instances. To account for asset serviceability, we generate a set of assets of varying type including physical assets, autonomous assets and virtual assets. For the autonomous assets we utilize the generated ALFUS scores in order to specify varying levels of capability and constraints. We also specify an asset worth, to denote it's value to a particular owner. For autonomous assets, the worth is set dynamically based on it's autonomous capability. In order to represent the fluid notion of assets dynamically coming online/offline during a mission and to enable asset utilization requests to be modeled we define the concept of a \emph{live asset inventory} that stores the real-time status of a collection of assets for every coalition partner on every mission instance. We then assign a random set of assets to each coalition partner's inventory, where the number of assets to assign is user configurable. Also, we generate a random starting configuration for each asset including a latitude/longitude location, a risk of adversarial compromise and a Boolean value indicating if this asset is available to use. The bounding box used to generate the initial asset locations is also user configurable. Finally, we create the concept of an asset request which indicates that a coalition partner requested to use an asset from another coalition partner on a given mission instance at a particular time. The request is also annotated through user configurable logic to determine an \emph{approve} or \emph{reject} decision which outlines whether this request is successful or not. 

\begin{figure*}[h!]
\centering
\includegraphics[width=\columnwidth*2]{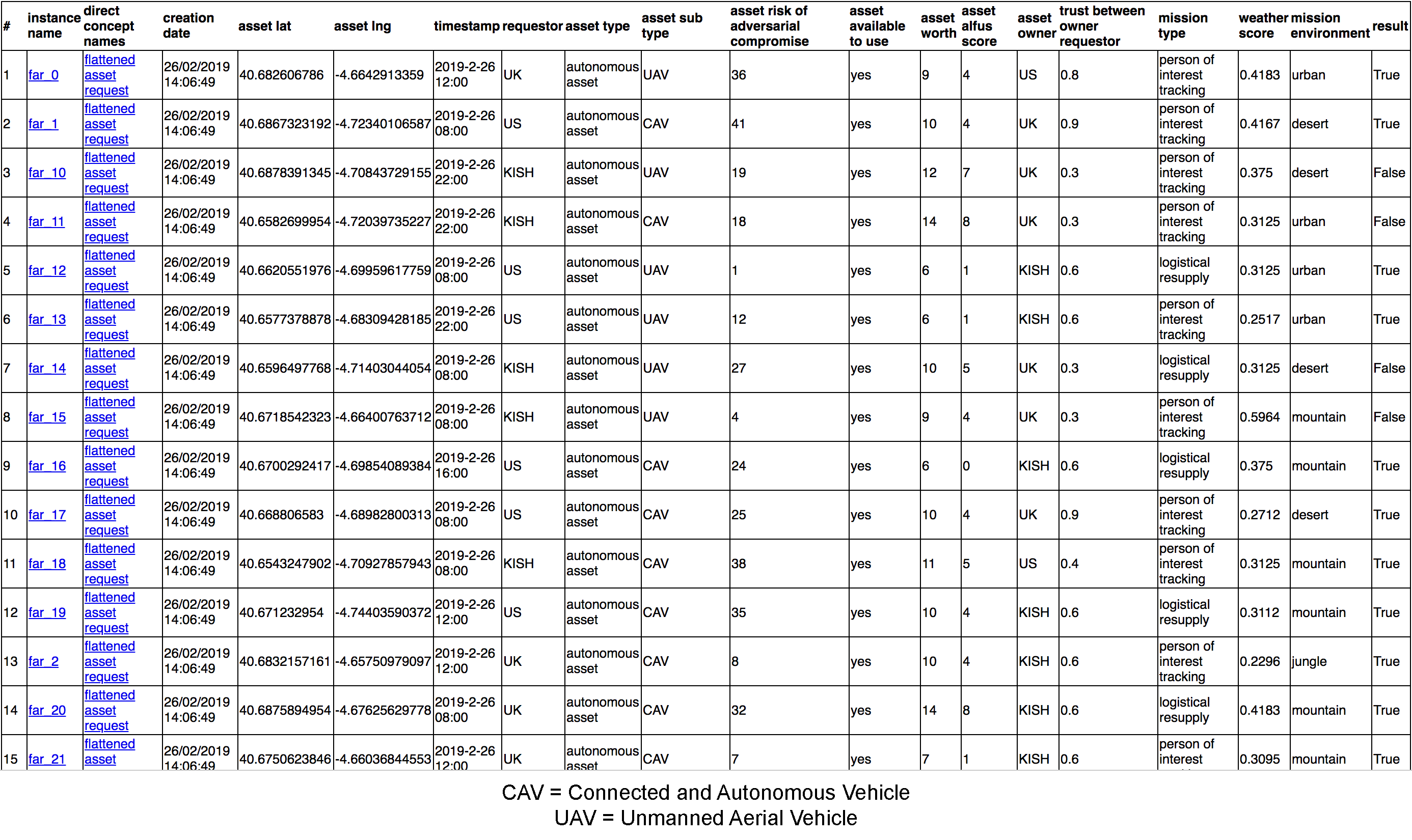}
\caption{Generated Asset Requests using logic defined in Figure~\ref{fig:customized-logic}}
\label{fig:generated_requests}
\end{figure*}

\begin{figure}
    \centering
    {\footnotesize{
\begin{lstlisting}
{
  "trust": {
    "comparison": "gt", 
    "value": 0.3
  }, 
  "asset": { 
    "available to use": { "eq": "yes" },
    "risk of adversarial compromise": { "lt": 40 }
  },
  "mission environment": {"eq": "urban|mountain" }
  "environmental condition instance": {
    "wind speed": {"lt": 30}
  }
}

\end{lstlisting}}}
    \caption{Customized logic to decide approve/reject for asset requests}
    \label{fig:customized-logic}
\end{figure}

\subsection{Customized Data Generation}
In order to generate customized data that supports a desired level of complexity, scripts within the online repository\footnotemark[\getrefnumber{dais}] can be used. These scripts accept user-defined logic to decide the final asset request annotations. Using the JavaScript Object Notation (JSON) format, logic can be specified as shown in Figure~\ref{fig:customized-logic} to decide whether to approve a request from one coalition partner to use an asset from another coalition partner. For example, Figure~\ref{fig:customized-logic} states that if the coalition partner who owns the asset trusts the coalition partner requesting the asset by a trust value $\geq 0.3$ and the asset is currently available to use, the risk of adversarial compromise of this asset is less than 40\%, the mission environment is either urban or mountainous and the wind speed is less than 30mph then approve this asset request, otherwise the asset request should be rejected. \ac{ce} facts will then be inserted into the knowledge base to define each annotated request. A sample of generated asset requests is shown in Figure~\ref{fig:generated_requests}.

This enables researchers to specify custom, potentially complex logic to define a policy model that can be used to evaluate future \ac{gpbm} architectures.
\section{Conclusion}\label{sec:conclusion}
This paper has presented a conceptual domain model and dataset for use on the \ac{dais} research programme. The domain model has been built using an agile representation framework which enables future work to extend our model easily using controlled natural language. Also, the dataset complexity can be controlled using custom logic declarations that will support future evaluations for \acp{gpbm}. Future work involves the ability to support partial data generation to simulate a real-world scenario where a complete dataset is not available and to investigate the suitability of Generative Adversarial Neural Networks for the suitability of synthetic data generation in the coalition environment.
\section*{Acknowledgement}
This research was sponsored by the U.S. Army Research Laboratory and the U.K. Ministry of Defence under Agreement Number W911NF-16-3-0001. The views and conclusions contained in this document are those of the authors and should not be interpreted as representing the official policies, either expressed or implied, of the U.S. Army Research Laboratory, the U.S. Government, the U.K. Ministry of Defence or the U.K. Government. The U.S. and U.K. Governments are authorized to reproduce and distribute reprints for Government purposes notwithstanding any copyright notation hereon.

\balance
\bibliographystyle{IEEEtranN}
\bibliography{ref}

\end{document}